\documentclass[runningheads]{llncs}

\usepackage[mobile,year=2024,ID=991]{eccv}
\usepackage{eccvabbrv}

\usepackage{graphicx}
\usepackage{booktabs}
\usepackage{graphicx}
\usepackage{amsmath}
\usepackage{amssymb}
\usepackage{tikz}
\usepackage{multirow}
\usepackage{ragged2e}
\usepackage{wrapfig}
\usepackage{ulem}
\usepackage{marvosym}
\def \names {SEK }

\newcommand{\revise}{}
\newcommand{\circledplus}{%
    \mathbin{% Ensures proper spacing
        \tikz[baseline=(char.base),yshift=0.2em]{
            \node (char) {$c$};
            \draw (char.center) circle (0.35em);
            \draw (char.center) -- ++(45:0.1em) -- ++(-135:0.2em);
        }%
    }%
}
\usepackage[accsupp]{axessibility}  % Improves PDF readability for those with disabilities.

\usepackage{hyperref}

\usepackage{orcidlink}
\usepackage{ulem}

\begin{document}

\title{External Knowledge Enhanced 3D Scene Generation from Sketch} 


\author{Zijie Wu\inst{1*}\orcidlink{0000-0002-0675-6525}  \and
Mingtao Feng\textsuperscript{2\Letter}\orcidlink{0000-0003-0384-3743}   \and
Yaonan Wang\inst{1}\orcidlink{0000-0002-0519-6458}\and\\
He Xie\inst{1} \orcidlink{0000-0003-3736-5508}\and 
Weisheng Dong\inst{2}\orcidlink{0000-0002-9632-985X}\and
Bo Miao\inst{3}\orcidlink{0000-0002-3025-4429}\and
Ajmal Mian\inst{3}\orcidlink{0000-0002-5206-3842}}

%Zijie Wu,Mingtao Feng,Yaonan Wang,He Xie,Weisheng Dong,Bo Miao,Ajmal Mian

\authorrunning{Zijie Wu et al.}

\institute{Hunan University\and Xidian University \and University of Western Australia }
%\institute{\inst{1}Hunan University }

\maketitle

% \begin{abstract}
%   The abstract should summarize the contents of the paper. 
%   LNCS guidelines indicate it should be at least 70 and at most 150 words.
%   Please include keywords as in the example below. 
%   This is required for papers in LNCS proceedings.
%   \keywords{First keyword \and Second keyword \and Third keyword}
% \end{abstract}
\makeatletter{\renewcommand*{\@makefnmark}{}
\footnotetext{$^*$~~~~Work performed during visit at the University of Western Australia \\
\Letter  \quad Corresponding author.}}

\begin{abstract}
\vspace{-3mm}
Generating realistic 3D scenes is challenging due to the complexity of room layouts and object geometries.
We propose a sketch based knowledge enhanced diffusion architecture (SEK) for generating customized, diverse, and plausible 3D scenes. SEK conditions the denoising process with a hand-drawn sketch of the target scene and cues from an object relationship knowledge base. We first construct an external knowledge base containing object relationships and then leverage knowledge enhanced graph reasoning to assist our model in understanding hand-drawn sketches. 
A scene is represented as a combination of 3D objects and their relationships, and then incrementally diffused to reach a Gaussian distribution.
We propose a 3D denoising scene transformer that learns to reverse the diffusion process, conditioned by a hand-drawn sketch along with knowledge cues, to regressively generate the scene including the 3D object instances as well as their layout. 
Experiments on the 3D-FRONT dataset show that our model improves FID, CKL by 17.41\%, 37.18\% in 3D scene generation and FID, KID by 19.12\%, 20.06\% in 3D scene completion compared to the nearest competitor DiffuScene.
\end{abstract}

\vspace{-9mm}
\section{Introduction}
\vspace{-1mm}
There is an increasing demand for tools that automate the creation of artificial 3D environments for applications in game development, movies, augmented/virtual reality, and interior design. 
Sketch based 3D scene generation allows users to control the generated scene entities through a rough hand-drawn sketch. 
Several methods for 3D scene generation rely on an input image~\cite{fastsynth,nie2023learning} to guide the generation process for alignment with the input.
However, such methods focus on leveraging 2D-3D consistency for supervision, which restricts diversity in the generated scene.
Moreover, obtaining an image that serves as a 2D rendering of the intended 3D scene is not always straightforward.

\begin{figure}[!t] 
		\centering
		\begin{tikzpicture}[inner sep=0pt,outer sep=0pt]
		
		\node[anchor=south west] (A) at (0in,0in)
		{\includegraphics[width=0.92\textwidth,clip=false]{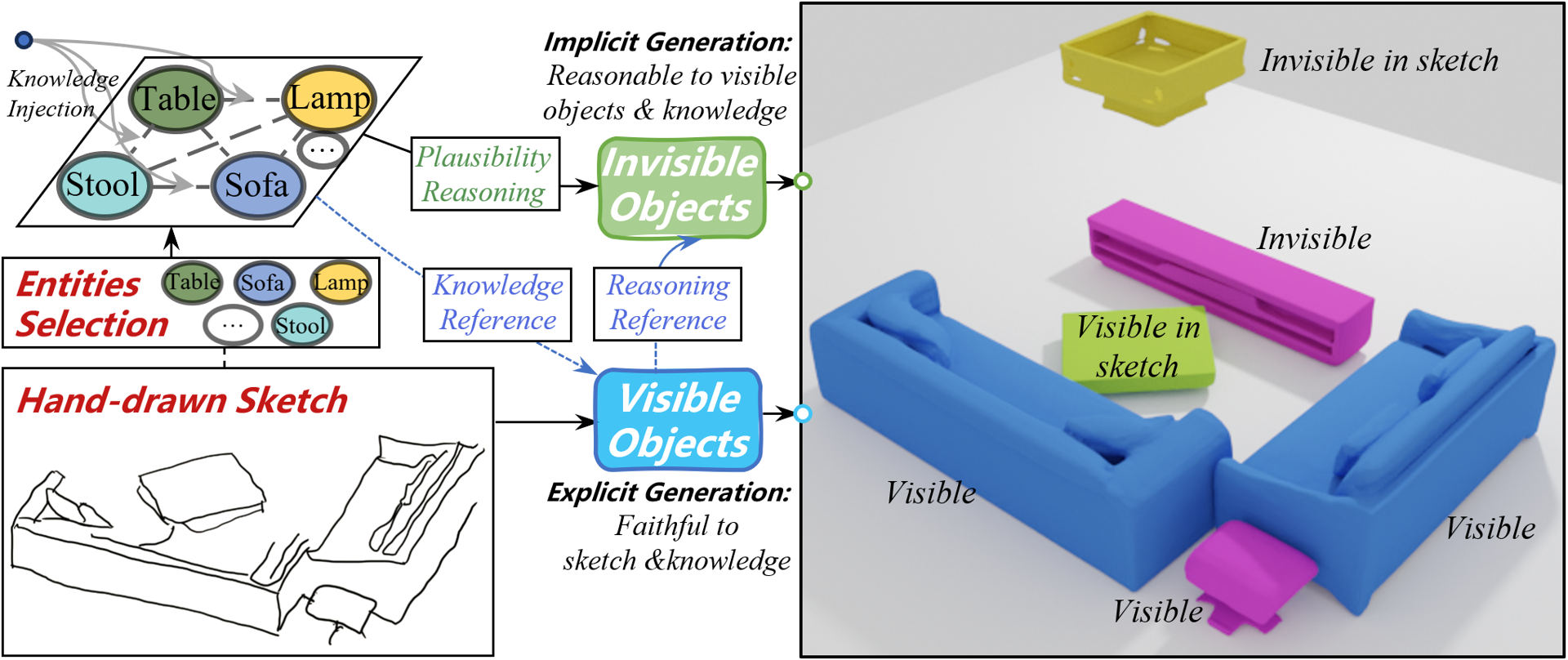}};		
		\end{tikzpicture}	
  \vspace{-3.5mm}
		\caption{\label{fig:overall} Our method generates a 3D scene from an input sketch and entities, enhanced by external knowledge. It follows explicit visual cues in the sketch for \textit{visible} objects along with their relationships and employs plausibility reasoning to add objects that are not explicitly depicted (\textit{invisible}) in the sketch, to generate a coherent scene.}
		\vspace{-7mm}
\end{figure}

Recently, sketch based methods~\cite{sketch-a-shape,sketch-text-shape,sketch2model,sketch3} have been proposed for user-specified 3D modeling. However, these methods focus primarily on  generating single 3D objects. %the geometric properties of objects. 
While much progress has been made in generating high-quality 3D objects, generation of complete 3D scenes is still challenging given the complex scene layouts, diverse object geometries and strong coherence between objects. 
For example, a chair can be placed underneath a table or around it, or may even be to the side of a bed. 
Various arrangements are possible for a chair and its neighboring objects, and each one must follow some rules including object-object relationships and space occupancy. 
To improve the representation and comprehension of 3D scenes, external knowledge 
has been introduced for multiple primary tasks, e.g., scene graph generation~\cite{feng20233d-scene-graph1}, robotic grounding~\cite{gao2021room-navigator1-robotic-gd1}, visual question answering~\cite{Parelli_2023_CVPR_vqa1knowledge}, and semantic segmentation~\cite{hou2022point-segmentation-knowledge1}. 
This involves reusing ontologies and integrating existing knowledge for improved outcomes.
External knowledge has been a prevailing technique in transferring implicit representations between scenes for improved performance in various 3D vision tasks. 
In this paper, we leverage external knowledge to provide auxiliary information for completing implicit scene patterns, that are not obvious in the sparse ambiguous hand-drawn sketch, and guide our proposed indoor 3D scene generation model. %, which makes learning more challenging. 

Existing methods use simple hand-crafted object relationships for generating 3D scenes. For instance, GRAINS~\cite{li2019grains-external_knowledge3} organize the scene objects into simple scene graph hierarchies that are manually defined. 
Furthermore, numerous works generate indoor scene layouts~\cite{paschalidou2021atiss,3dsln,fastsynth} in the form of object identities and bounding boxes and then \textit{retrieve} existing furniture shapes from a repository for placement inside those bounding boxes. Hence, the generated layouts as well as the object shapes both lack diversity.

We propose a 3D scene generation method (see Fig.~\ref{fig:overall}) 
that creates custom, diverse and plausible 3D scenes from hand-drawn sketches and entities, enhanced by external knowledge of object relationships. 
Our method takes a sketch as the main scene description and leverages external knowledge cues to reduce ambiguity in inferring \textit{visible} objects (shapes and layout) in the sketch and enhance the generation diversity by including \textit{invisible} objects that are not drawn in the sketch. 
We build an external knowledge base to contain rich knowledge priors of relationships.
\textit{Invisible} objects are inferred from the knowledge base across the \textit{invisible} and \textit{visible} objects to maintain diversity, plausibility and alignment with user specifications. Based on the sketch and knowledge reasoning, the proposed conditional scene diffusion \textit{simultaneously} generates a 3D scene layout with detailed object geometries (see Fig.~\ref{fig:vis_diffusion}) with plausible structure and coherence among objects. 
Our contributions are summarized below:

\vspace{-1mm}
\begin{itemize}

\vspace{-1mm}
\item We propose an end-to-end generative model (SEK) to \textit{simultaneously} generate realistic 3D room layouts and object shapes based on hand-drawn sketches and object entities, enhanced by external knowledge.

\item We construct an external knowledge base that defines various object relationships, and serves as a foundational entity-relationship prior to provide additional guidance to the inference process. This improves the plausibility of the generated scenes, including layout and object shapes. 

\item We learn novel reasoning from external knowledge cues and hand-drawn sketches
to extract a relationship subgraph of the specified entities during inference and integrate it with sketch features to form the diffusion condition.

\item We propose a 3D denoising scene transformer that operates in the latent space and  converts the denoising 
scene representation into the frequency domain to alleviate 
the influence of constants corresponding to invalid objects (padded zeros in scene representation) that are added to make the number of objects per scene constant.

\end{itemize}

\vspace{-4mm}
\section{Related Works}
\vspace{-1mm}

\noindent {\bf Sketch based 3D Object Generation:} Sketches have been used as a sparse representation of natural images and 3D shapes~\cite{sketch2model2tv,sketch2model} as they are quite illustrative, despite their simplicity and abstract nature. %Existing sketch based methods focus on object level generation. 
Some works~\cite{lun20173d-sketch1} estimate depth and 3D normals from a set of viewpoints for an input sketch, which are then integrated to form a 3D point cloud. Others~\cite{jin2020contour-sketch2} represent the 3D shape and its occluding contours in a joint VAE latent space during training, enabling them to retrieve a sketch during inference to generate a 3D shape. Recently, Kong~\textit{et al.}~\cite{kong2022diffusion-sketch3} trained a diffusion model conditioned on sketches using multi-stage training and fine-tuning. Sanghi~\textit{et al.}~\cite{sketch-a-shape} used local semantic features from a frozen large
pre-trained image encoder, such as CLIP, to map the sketch into a latent space for diverse shape generation. These methods mainly focus on object level generation, which prefers simple shape information instead of hierarchical relationship generation in scenes. However, 3D scenes contain rich information, including different furniture types, object geometries, room layouts, etc, presenting significant challenges to the generation process. 

\noindent {\bf Knowledge Graphs in 3D scenes:} Prior knowledge has proven to be an effective source of information to enhance object and relation recognition~\cite{feng2023exploring}. Pioneering works, including ConceptNet~\cite{speer2017conceptnet}, VisualGenome~\cite{krishna2017visualgenome}, DBPedia~\cite{auer2007dbpedia}, and WordNet~\cite{miller1995wordnet}, have extensively studied the acquisition of label-pairs frequency as a primary source of relations. These methods have achieved great success in many applications such as image generation~\cite{johnson2018image-scenegraph1,yang2022diffusionimage-scenegraph2}, visual question answering~\cite{teney2017graph-vqa1,ding2022mukea-vqa2}, camera localization~\cite{armeni20193d-camera-1}, and robotic grounding~\cite{gao2021room-navigator1-robotic-gd1,rosinol2021kimera-robotic-gd2,hughes2022hydra-robotic-gd3}. 
Nevertheless, the integrated knowledge is not very useful in isolation since it is hard-coded in the form of intrinsic parameters. Hence, some methods bring external knowledge bases into 3D tasks related inference. Gu~\textit{et al.}~\cite{gu2019scene-external_knowledge1} extracted knowledge triplets from the ConceptNet knowledge bases to help scene graph generation. GBNet~\cite{zareian2020bridging-external_knowledge2} adopted auxiliary edges as bridges that facilitates message passing between knowledge graph and scene graph.  Li~\textit{et al.}~\cite{li2019grains-external_knowledge3} introduced hand-crafted relationships for 3D scene generation in a recursive manner. The complete scene is encoded as multiple properties, including geometry and relationship, and is then recovered in a suggested pattern. 
Although previous studies have taken notice of knowledge in the 3D area, they only implicitly mine the extra knowledge base or define the relationship pairs to strengthen the iterative scene recovery between relationships and objects while ignoring the intrinsic properties of the data for specific 3D scene knowledge representation.

\noindent {\bf 3D Scene Generation:} Early methods for 3D scene generation are based on GANs~\cite{yang2021indoor-GANs,li2023deep-vae2}, VAEs~\cite{Chattopadhyay_2023_WACV-vae1,li2023deep-vae2,purkait2020sg-vae3}, and Autoregressive models~\cite{wang2021sceneformer,gao2023scenehgn-autorg1,paschalidou2021atiss}. They are renowned for their ability to generate high-quality results quickly, yet they often face challenges of limited diversity and difficulties in producing samples that align well with user specifications. 
Numerous methods learn to produce faithful results under different input conditions, such as images~\cite{nie2023learning,tulsiani2018factoring-image1,nie2020total3dunderstanding-image2}, text~\cite{ma2018language-text1,tang2023diffuscene}, sketches~\cite{xu2013sketch2scene}, and wall layouts~\cite{gao2023scenehgn-autorg1,paschalidou2021atiss,jyothi2019layoutvae-layout1}.
Another approach in 3D scene generation is based on graph conditioning~\cite{3dsln,graph23d}. Graph-to-3D~\cite{graph23d} jointly optimizes models to learn both scene layouts and shapes conditioned on a scene graph. However, the scene graph does not directly reveal the relationships among objects. This necessitates a complete graph description, impacting their realism and applicability. %limiting its diversity.
Conditional 3D scene synthesis methods offer faithful scene recovery tailored to user specifications, while generative methods strike a balance between diversity and alignment with user specifications.

\vspace{-2mm}
\section{Diffusion Model for Scene Generation}
\begin{figure*}[!t] 
		
		\centering
        \includegraphics[width=0.99\textwidth,clip=false]{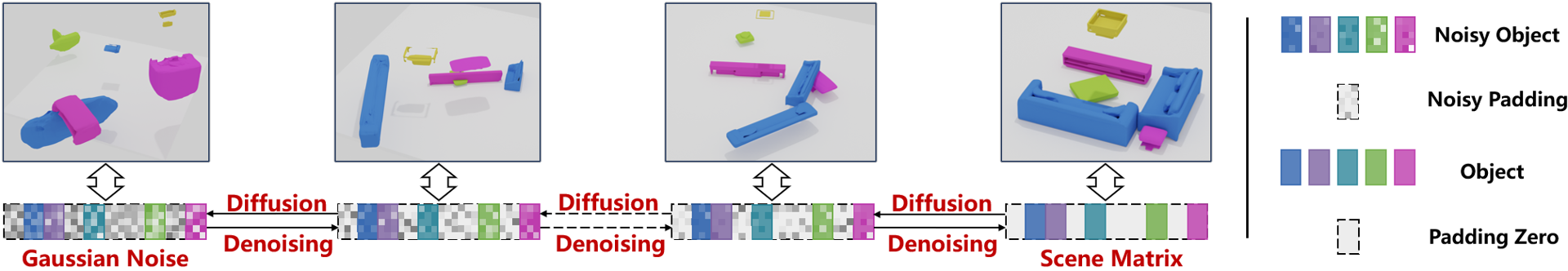}
        \vspace{-3mm}
        \caption{\label{fig:vis_diffusion} Demonstration of the scene diffusion/denoising process of the matrix field and the spatial field. The denoising process 
        samples from a Gaussian distribution and progressively denoises the sample for plausible and realistic scene generation. Note how the layout and 3D shapes are both \textit{simultaneously} denoised.} %TBD::
        \vspace{-7mm}
	\end{figure*}

\vspace{-1mm}
In the diffusion process, the data distribution is gradually destroyed into Gaussian noise following the Markov forward chain. A denoising process then recovers data from the Gaussian distribution with an iterative reverse chain. 
\revise{To devise our scene diffusion model for generating 3D scenes, we introduce matrix conversion to represent an indoor scene in the form of a matrix. All processes operate on the matrix field. Fig.~\ref{fig:vis_diffusion} illustrates how the diffusion and denoising processes mutually transform the Gaussian and target data distributions.}

\noindent\revise{\textbf{Matrix conversion} is proposed to encode the scene objects into parameters that specify their locations and shape attributes.} Given an indoor scene $\mathcal{O}$ containing a set of objects $\{\mathbf{o}_i\} |_{i=1}^{K}$, each object $o_i$ is characterized by a 1-D vector which concatenates its spatial location and latent shape attributes, $i.e.$ $\mathbf{o}_i = [\mathcal{G}_i,\mathcal{F}_i]\in \mathbb{R}^{D\times 1}$. 
Every 3D scene is normalized by relocating it in a world coordinate system where the floor center is the origin.

The placement location of each object $\mathcal{G}_i = [\mathbf{\alpha}_i, \mathbf{s}_i, \mathbf{t}_i]$ is defined by its axis-aligned 3D bounding box size $\mathbf{s}\in \mathbb{R}^{3\times 1}$, translation $\mathbf{t}\in \mathbb{R}^{3\times 1}$ and yaw angle $\alpha \in \mathbb{R}^{2\times 1}$. 
Following~\cite{yin2021center}, the yaw angle is parameterized as a 2D vector of sine and cosine values.
All objects are normalized to $[-1,1]^3$ first, and subsequently encoded into latent space. The shape latent $\mathcal{F}\in \mathbb{R}^\mathcal{F}$ is trained on DeepSDF~\cite{park2019deepsdf} to obtain a unique code per object. 

Since the number of objects can vary across scenes, we pad zero-vectors $\{\mathbf{z}_i| i\in(K+1,M)\}\in \mathbb{R}^{D\times 1}$ so that all scenes have a fixed number of $M$ objects. Here, $K<M$. 
All objects are concatenated to form a full scene representation: $\mathcal{O}\in \mathbb{R}^{D \times M}$.
\revise{%Following the matrix conversion, 
A scene is encoded as a unique matrix, where each row corresponds to an object with shape, location and size attributes. By synthesizing various combinations of object parameters, we can generate diverse scenes.}

\noindent\textbf{Diffusion Process}: In the forward chain of scene diffusion, the original scene matrix $\mathcal{O}_0 \sim q(\mathcal{O}_0)$ is gradually corrupted into a pre-defined $T$-step noised scene distribution following the Markov chain assumption until the Gaussian distribution is reached. Based on the Markov property, the joint distribution $\mathcal{O}_{1: T}$ is straight derived from the original scene matrix $\mathcal{O}_0$:
\vspace{-3mm}
\begin{small}
\begin{equation}
q\left(\mathcal{O}_{0: T}\right)=q\left(\mathcal{O}_0\right) \prod_{t=1}^T q\left(\mathcal{O}_t \mid \mathcal{O}_{t-1}\right),\quad
q\left(\mathcal{O}_t \mid \mathcal{O}_{t-1}\right) =\mathcal{N}\left(\sqrt{1-\beta_t} \mathcal{O}_{t-1},~ \beta_t \mathbf{I}\right),
\vspace{-3mm}\end{equation}
\end{small}

\noindent where $\mathcal{N}\left( \mu,\sigma^2 \right)$ denotes a Gaussian distribution and $\beta_t$ is the known variance defined during the diffusion process. 

\noindent \textbf{Deniosing process:} Since the forward chain concludes with a Gaussian distribution, we apply the reverse chain starting from a standard Gaussian prior and ending with the desired scene representation $\mathcal{O}_0$:
\vspace{-1mm}
\begin{small}
\begin{equation} \label{eq:sample}
p_\theta\left(\mathcal{O}_{0: T}\right)=p\left(\mathcal{O}_T\right) \prod_{t=1}^T p_\theta\left(\mathcal{O}_{t-1} \mid \mathcal{O}_t\right),\quad
p_\theta\left(\mathcal{O}_{t-1} \mid \mathcal{O}_t\right) =\mathcal{N}\left(\mu_\theta\left(\mathcal{O}_t, t\right), \sigma_t^2 \mathbf{I}\right),
\vspace{-1mm}\end{equation}
\end{small}

\noindent where $p_\theta$ is the inference step using parameterized network of the proposed scene denoiser. \names is trained by minimizing the cross-entropy loss between two diffusion chains in relation to the sketch and knowledge enhanced conditional feature $c$, and by learning the scene denoiser parameters $\theta$:
\vspace{-2mm}
\begin{small}
\begin{equation} \label{eq:learning}
          \min_\theta \mathbb{E}_{\mathcal{O}_0\sim q(\mathcal{O}_0), \mathcal{O}_{1:T}\sim q(\mathcal{O}_{1:T})}[\sum_{t=1}^T\log p_\theta(\mathcal{O}_{t-1}|\mathcal{O}_t, c)].%\notag       
\vspace{-1mm}\end{equation}
\end{small}

Following~\cite{ho2020denoisingdiffusion1} to minimize Eq.~\ref{eq:learning}, \names learns to match each $q\left(\mathcal{O}_{t-1} \mid \mathcal{O}_t,\mathcal{O}_{0}\right)$ and $p_\theta\left(\mathcal{O}_{t-1} \mid \mathcal{O}_t\right)$ by estimating the noise $\mathbf{\epsilon}_\theta$ under the condition $(c,t,\mathcal{O}_t)$ to match the added noise $\mathbf{\epsilon}$ in the diffusion process:
\vspace{-1mm}
\begin{small}
\begin{equation} \label{eq:lsce}
\mathcal{L}_{\mathrm{sce}} =\mathbb{E}_{c,t,\epsilon,\mathcal{O}_0}\left[\left\|\epsilon-\epsilon_\theta\left(c,t,\mathcal{O}_t\right)\right\|^2\right], \epsilon\sim \mathcal{N}(0,\mathbf{I}). 
\vspace{-1mm}\end{equation}
\end{small}

Scene diffusion progressively generates the 3D scene using the reverse chain:

\vspace{-1mm}
\begin{small}
\begin{equation}\label{eq:generate}
    \mathcal{O}_{t-1}=1/{\sqrt{\alpha_t}}\left(\mathcal{O}_t-{1-\alpha_t}/{\sqrt{1-\bar{\alpha}_t}} \mathbf{\epsilon}_\theta(c,t,\mathcal{O}_t)\right)+\sqrt{\beta_t}\mathbf{\epsilon}, % \notag
\vspace{-1mm}\end{equation}
\end{small}

\noindent where $\alpha_t=1-\beta_t$, $\tilde{\alpha}= \prod_{s=1}^t\alpha_s$, and $\epsilon$ is the standard Gaussian noise.
\revise{Building upon a well-defined scene diffusion, the proposed model is theoretically capable of generating high-quality and diverse 3D scenes.
}

\vspace{-2mm}
\section{Knowledge Enhanced Sketch based Guidance}% for Scene Diffusion Model}
\begin{figure*}[!t] 
		
		\begin{tikzpicture}[inner sep=0pt,outer sep=0pt]
		\centering
		\node[anchor=south west] (A) at (0in,0in)
		{\includegraphics[width=\textwidth,clip=false]{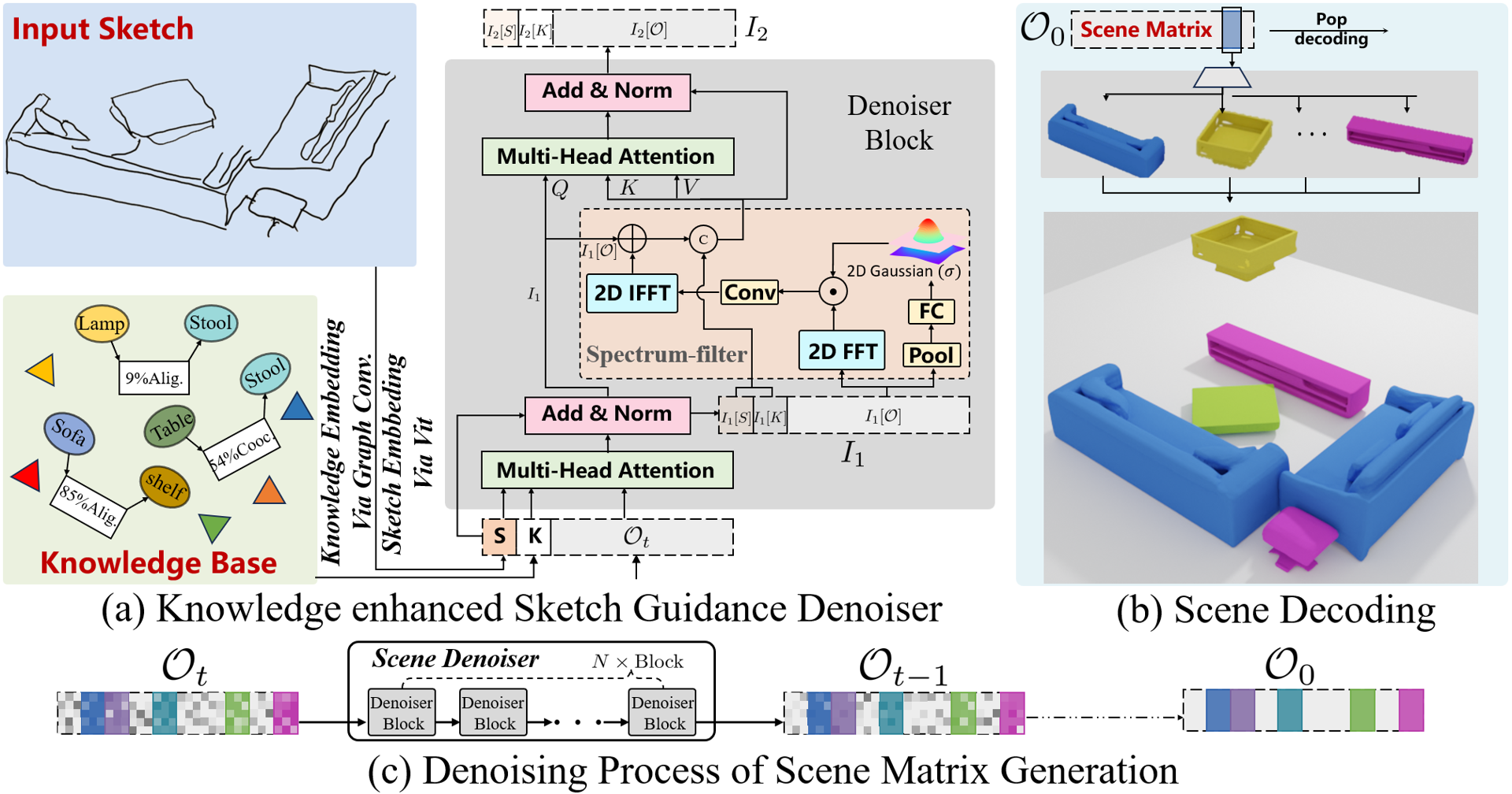}};

		\end{tikzpicture}

\vspace{-3mm}		\caption{\label{fig:framework} Proposed \names framework. (a) Sketch features are extracted by ViT and integrated with knowledge-enhanced reasoning features to form the denoising condition. The proposed 3D scene denoiser simultaneously generates plausible layouts and realistic 3D shapes in the matrix field.
(b) The generated scene matrix is decoded to form the complete scene.
(c) The denoising process: The scene denoiser starts from random noise and iteratively generates the scene matrix.}
		\vspace{-7mm}
	\end{figure*}

\vspace{-1mm}
\revise{Equipped with the scene diffusion model, the problem we need to address is how to ensure that the generated scene aligns with user description while preserving diversity, quality and plausibility. 
The input modality requires capturing the essence of the scene, providing a comprehensive description of the backbone while allowing flexibility and ambiguity, without confining to specific details, to foster diversity in generation.
To this end, we deploy a model with \textit{sketch} conditioning, offering strong flexibility, user-friendliness and diversity. Since all desired objects are not necessarily drawn in the sketch, \textit{object entities} are also provided as input to complement the sketch and extract cues from the knowledge  base.
As depicted in Fig.\ref{fig:framework}(a), the knowledge-enhanced sketch is integrated via multi-head attention. We employ a spectrum-filter (SF) to enhance meaningful object features. The conditional denoising process iteratively predicts noise for scene matrix generation (Fig.\ref{fig:framework}(c)). Once the scene matrix is generated, the complete scene is decoded into a spatial field through data pop-out decoding along the object number dimension (Fig.~\ref{fig:framework}(b)). Meaningless padding zeros are discarded.}

\revise{Sketch serves as our primary medium for convenient user interaction, however, it lacks details to provide precise instructions for the scene generation. This aligns well with our objective of allowing diversity in generation while remaining faithful to user specifications. To enhance the instructions contained in a sketch, we integrate external knowledge that clarifies vague information. For example, when faced with a sparse sketch depicting either a "\textit{table-aligned-sofa}" or a "\textit{stool-aligned-sofa}," querying knowledge can assist in determining which scenario is more likely to occur in indoor scenes.}

\revise{Knowledge serves as a complement or extension to the sketch in our framework. For \textit{visible} objects in the sketch, knowledge facilitates bidirectional validation to complement sketch descriptions. When a user provides object entities that are not visible in the sketch, knowledge helps the model to accommodate plausible object shapes based on the visible side of the relationships. 
In summary, knowledge enhancement provides several advantages: 1) Visible sketch enhancement: When object relations are depicted, knowledge can complement any ambiguous object descriptions in the hand-drawn sketch to enhance the plausibility of the generation. 2) Invisible description complement: If the depicted relations in the sketch do not contain the desired object entities, knowledge can accommodate the \textit{invisible} objects  by relating them to the depicted ones. For example, placing an appropriate "table" alongside a "sofa".
}

\vspace{-3mm}
\subsection{\label{ku1} Knowledge Base}
\vspace{-1mm}%Initialization}
\revise{Generating a complete scene from a sketch often encounters challenges due to vague description and insufficient details.
The proposed knowledge base helps in keeping the generated scenes realistic.}
We view knowledge as an essential complement to the scene sketch, effectively addressing its inherent sparsity and ambiguity. \revise{This section introduces our external knowledge base designed to retain extensive %knowledge of relationships for prior
relationship priors for injection into the inference process.  Knowledge of object relationships is first extracted from this external knowledge base and then dynamically learned to correspond with the sketch, facilitating interaction between the desired object entities.} %

\revise{We define knowledge base $KB=(\mathcal{V}, \mathcal{R},p)$ as a repository containing a set of triplet relations where $\mathcal{V}$, $\mathcal{R}$, and $p$ denote object nodes, their inner relation edges, and their edge probability respectively. The nodes $\mathcal{V}$ consist of a set of object types $f = \{f_1,f_2,\cdots,f_N\}$ and the relation edges $\mathcal{R}$ contain multiple predefined relations.
%which formulated as ``object\_A-Relation (probability)-object\_B''.
Each triplet indicates the probability of the given relation existing between objects. This information is considered as external knowledge to reveal object relationships and then enhance the sketch descriptions in our framework.
The predefined edge relationships $\mathcal{R}$, which include knowledge cues, are initially extracted from all indoor scenes in the given dataset. These relationships are then normalized and stored in a unique knowledge graph, forming the designed external knowledge base. Here, we introduce our scheme details to extract the priors of object relations, which comprise the external knowledge base $KB$.}

\vspace{1mm}
\noindent\revise{\textbf{Object Relationship Construction}:} External knowledge integration aids in generating reasonable semantic entities and their relationships. Hence, we extract multiple knowledge relations $\mathcal{R}$ from the indoor scenes dateset. For generating this knowledge relation, each scene is divided into smaller functional groups using the density-based clustering algorithm, DBSCAN~\cite{campello2020density_dbscan}. Initially, each object in the scene is voxelized, and then clustered into groups.

Overall, for a 3D scene $S$, we cluster it as $S = \{\left\langle \mathbf{g}_1, \mathbf{g}_2,\dots,\mathbf{g}_K\right\rangle \| \mathbf{g}_i= (o_i, \dots, o_j)\}$, where $\mathbf{g}_i$ represents the $i^{th}$ clustered group comprising variable number of objects. Inspired by ~\cite{3dssg,li2019grains-external_knowledge3,gao2023scenehgn-autorg1}, the extracted relationships between paired objects are categorized as \revise{adjacent} relations within the same group $\mathbf{g}_i,\mathbf{g}_j, i=j$:
\vspace{-2mm}
\begin{equation}\label{adj}
\mathcal{R}_a=\{\textit{``Attachment''},\textit{``Alignment'',``Dependent''}\}
\vspace{-2mm}\end{equation}
 and \revise{distant} relations cross different groups $\mathbf{g}_i,\mathbf{g}_j,i\neq j$:
 \vspace{-2mm}\begin{equation}\label{dist}
\mathcal{R}_d=\{\textit{``Co-occurrence'',``Parallel Collinearity''}\}
\vspace{-2mm}\end{equation}
Unlike previous works that focus on object-wise relationships, we  extract multiple relations to depict how entities are organized. These relations are defined as follows: (a) \textit{Attachment}: The minimum distance between two adjacent 
 object voxels is smaller than voxel length. (b) \textit{Alignment}: Any plane of bounding boxes from two adjacent objects is coplanar. (c) \textit{Dependent}: Adjacent pairwise relations in the same group that do not belong to attachment and alignment. (d) \textit{Parallel Collinearity}: Horizontal axes of bounding boxes of the objects of different groups are parallel. (e) \textit{Co-occurrence}: Identifying two objects that co-occur in the same scene of different groups.

\noindent\revise{\textbf{Relationship Probability Counting:}} 
We count the number $n_{ij}$ of these relationships using a clustered scene structure derived from the 3D-FRONT Dataset~\cite{fu20213d} to perform knowledge initialization. $i,j$ denote indices of the entities, such as $chair,~desk$. Each probability $p$ is normalized by: %$p_{ij} = 1/{1+e^{-10\cdot n_{ij}^\mathcal{R}/max(n)^\mathcal{R}}}$.
%$p_{ij} = 1/{1+e^{-10\cdot n_{ij}^\mathcal{R}/max(n)^\mathcal{R}}}$.
%Each weight is normalized as:
\vspace{-2mm}
\begin{equation}
    \label{eq:normas} p_{ij} = 1/{1+e^{-10\cdot n_{ij}^\mathcal{R}/max(n)^\mathcal{R}}}.
\vspace{-2mm}\end{equation}
%\vspace{-2mm}

The defined relationship edges $\mathcal{R}$ and their corresponding probability $p$ are then integrated into the built-in knowledge base $KB$. 
The proposed \names learns the object correlations among the given objects types, enabling it to dynamically interact with the knowledge base and assist in the diffusion module for scene generation. We build the knowledge base from 5,754 scenes in the  dataset. More details are in the supplementary material.

\vspace{-2mm}
\subsection{\label{ku2} Knowledge-enhanced Graph Reasoning}
\vspace{-1mm}
\revise{The constructed knowledge base effectively reveals the potential spatial relationships among various pieces of objects.}
Based on the desired object types (\textit{chair, table, desk, etc.}), we propose a knowledge-enhanced graph reasoning module (KeGR) to incorporate external knowledge from the initialized knowledge base for comprehensive room generation reasoning. For object types $\{f_i\}^n_{i=1}$ that the scene demands, we initialize each object representation $h_i$ of $f_i$ via GloVe so that $h_i\in \mathbb{R}^{1\times D_\omega}$. \revise{Next, we obtain a subset of $\{f_i\}^n_{i=1}$ to construct a fully connected subgraph $G_{i}^E=$ $(h_{i}^E, \mathcal{E}_{i}^E, \mathcal{P}_{i}^E)\in \{KB\}$. $h_{i}^E \in \mathbb{R}^{n\times D_\omega}$ denoting its node feature matrix}. We represent edge and probability $\mathcal{E}_{i}^E, \mathcal{P}_{i}^E$ as the initial adjacency matrix $A_{i}^E \in \mathbb{R}^{n\times n}$. Knowledge-enhanced graph reasoning is achieved via multistep graph convolutions:
\vspace{-2mm}
\begin{equation}\label{gr}
    H_i^{E(j)}=\delta\left(A_i^E H_i^{E(j-1)} W^{E(j)}\right),
\vspace{-3mm}\end{equation}
where $j$ denotes the $j^{th}$ step of graph reasoning and $\delta$ is the activation function. $W^{E(j)}$ is a learnable parameter, and $H^{E(j)}$ is the node feature matrix of $G^E$ at $j_{th}$ step. After $J$ iterations, we term $H_i^{E(j)}$ as the final node feature matrix of the graph reasoning of the current relationship. With all relation feature update, we get a final feature matrix group $\{H_i^{E(J)}\}|_{i=0}^{r}$, where $r$ is the index related to relationships (\ref{adj},\ref{dist}) of knowledge base. %: \{\textit{``Attachment''}, \textit{``Alignment''},\textit{``Dependent''},\textit{``Co-occurrence"''},\textit{``Parallel Collinearity''}\}.
We perform a $1 \times 1$ convolution across the relationship dimension to get the graph feature that is used to condition the scene diffusion:
%\vspace{-2mm}
\begin{equation}
    H^G = \delta(conv_{1\times 1}(H_0, H_1, \dots, H_r)).
\vspace{-3mm}\end{equation}

\vspace{-3mm}
\subsection{\label{ku3} Knowledge enhanced Sketch Guided Denoiser}
\vspace{-1mm}
\revise{The denoiser serves as the key module of the scene diffusion model. It predicts the noise $\epsilon_\theta$ for denoising process, thereby enabling the iterative generation of the 3D scene using Eq.~\ref{eq:generate}.}
With the external knowledge, our KeGR module produces a rich feature representation for guiding scene generation. Given the specified entities, very diverse scenes can possibly be generated. Hence, to guide the generation process w.r.t.~user alignment, we include a sketch as a complementary description.
Sketches are highly expressive, inherently capturing subjective and fine-grained visual cues. Furthermore, its advantage lies in the combination of easy access and vivid description. Conditioned on sketch based knowledge reasoning features, the proposed \names denoises the 3D scene from a random point in Gaussian distribution. We employ ViT~\cite{dosovitskiy2020vit} as our sketch embedding backbone to obtain the sketch condition $H^{S}$ maintaining the details. The conditional feature $c$ is formed as the concatenation $[H^{S},H^{G}]\in \mathbb{R}^{D \times 2}$ of sketch and graph features.

\vspace{1mm}
%\noindent\textbf{Component Enhancement filtering:}
In the forward chain of scene diffusion, the scene representation matrix $\mathcal{O}^{D\times M}$ (along with the padding) is diffused by adding Gaussian noise. % along with the padding values $\mathcal{O}^{D\times (M-K)}$. 
\revise{As depicted in Fig.~\ref{fig:vis_diffusion}, padding occupies a significant portion of the scene matrix, potentially overwhelming the object information as the noise level increases. Padding is introduced in scene representation to make its dimension fixed during training. However, during inference, there is no mask available to filter out any padded values that are generated but have no specific meaning and overwhelm the desired shapes. The absence of a padding mask and the unavailability of the number of generated objects make it difficult to filter out disturbance components effectively. 
To address this problem, we propose component enhancement through a spectrum-filter with the intention to filter out the padding, ensuring that the prediction receives sufficient information from the valid object components.}
Compared to the valid object representations, we observe that the padding zeros have a low-frequency variance distribution. Note that this low-frequency distribution vanishes as noise is systematically added to $\mathcal{O}_0$ step-by-step, following the Markov chain assumption, and finally reaches the single-kernel Gaussian distribution. We apply a high-pass filter to suppress the low-frequency padding in the spectral domain. Let $\mathcal{O}_I$ denote the output of the attention blocks; the proposed spectrum-filter is computed as
\vspace{-1mm}
\begin{small}
\begin{equation}
\mathrm{EF}(\mathcal{O}_I, B)=\mathcal{O}_I+ e^{-t} \Theta_{I F F T}\left(\operatorname{Conv}\left(\sigma( \mathcal{O}_I,B) \circledast  \Theta_{F F T}(\mathcal{O}_I)\right)\right), \notag \vspace{-1mm}
\end{equation}%\vspace{-3mm}
\end{small}

\noindent where $t$ is the time step and $\circledast$ denotes high-pass filtering with adaptive Gaussian smoothed filters $\sigma(\mathcal{O}_I,B)$ (with bandwidth $B$), which has the same spatial size as $\mathcal{O}_I$. $\Theta$ denotes the spectrum operation using Fourier transform. Following~\cite{miao2023spectrum}, we create an initial 2D Gaussian map based on bandwidth $B$ and apply the predefined weight parameter associated with time step $t$ to scale the filter.

Our spectrum-filter block enhances meaningful object features in the encoded scene representation. 
As shown in Fig.~\ref{fig:framework}(a), the scene transformer performs feature embedding first to get the embedding context initialization $\mathcal{I}=[c,t,\mathcal{O}_t]\in \mathbb{R}^{D\times (M+3)}$, where $t \in \mathbb{R}^{D\times 1}$ is the time step embedding and $\mathcal{O}_t \in \mathbb{R}^{D \times M}$ is the scene representation at time $t$. \revise{We begin by applying multihead attention at dimension $M$ to capture the relevance of each element to every other element in the sequence. For example, we explore the guidance correlation between sketches and knowledge, the relevance of guidance among different conditions and every piece of object, as well as the interactions among different pieces of objects:
\vspace{-2mm}
\begin{small}
\begin{equation}
       I_1\in \mathbb{R}^{D\times (M+3)}  = Atten_1 (Q_1 = K_1 = V_1 = \mathcal{I}).
\vspace{-2mm}\end{equation}
\end{small}
Next, we encode $\mathcal{I}_1$ using a transformer encoder to enrich its semantic information for each instance and then follow it by the proposed spectrum-filter block:
\begin{small}
    \vspace{-4mm}\begin{equation}
       I_2\in \mathbb{R}^{D\times (M+3)}  = Atten_2 (Q_2 = I_1,  K_2 = V_3 = I_1[S]\circledplus I_1[K] \circledplus EF(I_1[\mathcal{O}])),
\vspace{-2mm}\end{equation}
\end{small}
\noindent where, $\circledplus$ denotes concatenation.}
Finally, we sample
$\epsilon_{t-1}$ with dimensions congruent to those of scene $\mathcal{O}$ for prediction, using a set of regressive steps, where $\epsilon_{t-1}$ is the current predicted noise using the scene denoiser.% at time step $t-1$.

\revise{Overall, the 3D scene denoiser takes the context embedding $\mathcal{I}$ as input to perform spatial self-attention using the multi-head attention block. It is then fed to the spectrum-filter to enhance the features of valid objects and suppress invalid padding. Finally we take the output with dimension congruent to $\mathcal{O}_t$ as the predicted noise $\epsilon_{t-1}$ for supervision.}

\vspace{-2mm}
\section{Experiments}
\vspace{-1mm}
\noindent \textbf{Datasets:} We train and test our method on three downstream tasks: 3D scene generation, 3D scene completion, and knowledge transfer validation.
For the generation task, we use three types of indoor rooms from the 3D-FRONT dataset~\cite{fu20213d}, including 4041 $Bedrooms$, 900 $Dining rooms$, 813 $Living rooms$. We randomly split the data into training-test sets at 80-20\% ratio. \revise{To acquire sketch, we first render images from 21 views using BlenderProc from each 3D scene uniformly when the viewpoint axis is $z_{vp}>0$. We then apply Canny edge detection~\cite{Canny}
to the rendered scene images to acquire their edge sketches.} We manually remove the walls so that each scene contains hand-drawn looking sketches of the object. 
In scene completion, we randomly mask 30-80\% of the objects in the scene and render it to acquire the sketches with viewpoint the same as in generation task following the above rendering process. 
Finally, we test the effectiveness of knowledge transfer by transferring knowledge from the 3D-FRONT dataset to the ScanNet dataset~\cite{dai2017scannet}. ScanNet is a real indoor scene dataset with 1,513 rooms of 21 different types. Common categories between ScanNet and 3D-FRONT dataset are selected for our knowledge transfer experiment.
We retrieve objects of ScanNet from ShapeNet~\cite{chang2015shapenet} to acquire consistent objects across scenes to maintain the same setting as in 3D-FRONT. 

\noindent \textbf{Baselines:} \revise{We compare with state-of-the-art scene generation methods which can be categorised into retrieval-based and generation-based methods. In the former category, current works focus on the 3D scene layout generation followed by object placement using shape retrieval to form the complete scenes.
We select the major floor plan~\cite{fastsynth,wang2021sceneformer,paschalidou2021atiss}, room size~\cite{yang2021indoor}, and graph~\cite{3dsln} based scene generation methods. Besides, some unconditional generation methods~\cite{yang2021scene-syc2gen,tang2023diffuscene} are also included for comparison. In the latter category, methods generate both shape and layouts to directly form the 3D scenes. We select the graph~\cite{graph23d} based and unconditional~\cite{nie2023learning} generation methods for comparison. For a fair comparison, we ensure the training data of  baselines is the same and that each model has its required modality.} 
\revise{Furthermore, we also compare with the most relevant sketch based method, Sketch2Scene~\cite{xu2013sketch2scene}. The ATISS and Sceneformer are adopted to accept module plugin of our sketch condition for comparison.}

\begin{figure*}[!t] 
		
		\centering
       \includegraphics[width=0.98\textwidth,clip=false]{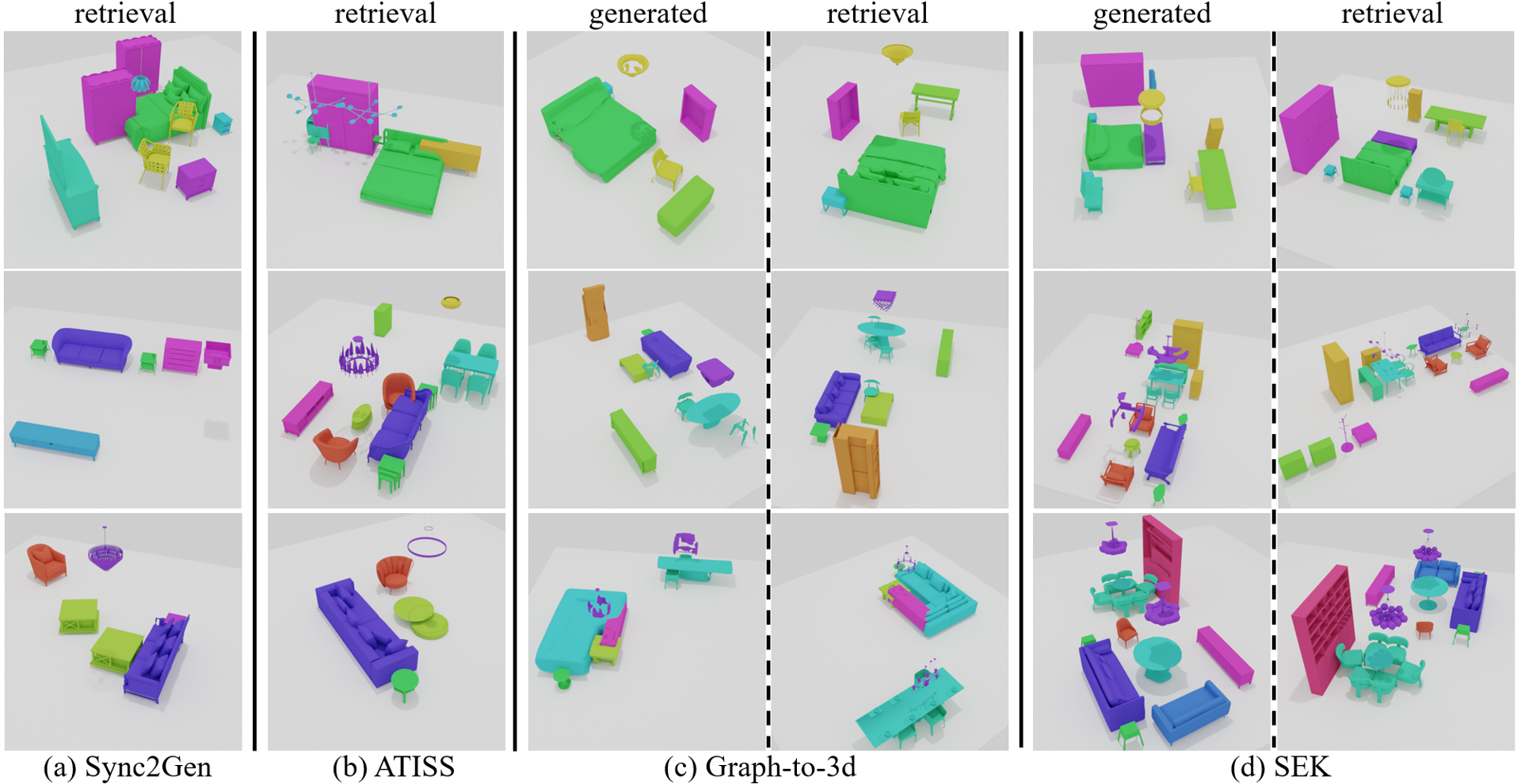}

		\vspace{-3mm}
		\caption{\label{fig:vis} Qualitative comparison. 
  %Visualization of realistic scene generation is presented. 
  Syn2Gen and ATISS perform retrieval using 3D bounding boxes. Graph-to-3d and our method perform generation but we also show the corresponding retrieval results by searching nearest neighbor of shape code for comparison. 
  Our method performs higher quality generation with detailed shapes and better plausibility of relationships.}
		\vspace{-7mm}
	\end{figure*}

\noindent \textbf{Evaluation Metrics:} Following previous works~\cite{wang2021sceneformer,yang2021scene-syc2gen,paschalidou2021atiss}, we use Frechet Inception Distance (FID), Kernel Inception Distance (KID $\times$ 0.001), Scene Classification Accuracy (SCA), and Category KL Divergence (CKL $\times$
0.01) to measure the plausibility and diversity of 1,000 generated scenes. Additional information regarding evaluation metrics can be found in the supplementary materials.

\begin{table*}[hbp]
\centering
\vspace{-6mm}
\caption{\label{tab:generation} Comparative results on the 3D-FRONT dataset. For Scene Classification Accuracy (SCA), a score closer to 50\% is better as it means that the generated distribution is closer to target distribution.} 
\vspace{-4mm}
%\footnotesize
\scriptsize
\addtolength{\tabcolsep}{2pt}
\begin{tabular}{l |c c c|c c c|c c c}
\hline \hline  \multirow{2}{*}{\text { Method }} & \multicolumn{3}{|c|}{\text { Bedroom }} & \multicolumn{3}{|c|}{\text { Dining room }} & \multicolumn{3}{|c}{\text { Living room }} \\ \cmidrule{2-10} 
& FID  $\downarrow$ & $\mathrm{SCA}$ \% & $\text {CKL } \downarrow$ & $\mathrm{KID} \downarrow$ & $\text {SCA } \%$ & $\mathrm{CKL} \downarrow$ & $\mathrm{KID} \downarrow$ & $\text {SCA } \%$ & $\text {CKL} \downarrow $\\ \hline
DepthGAN~\cite{yang2021indoor} & 40.15 & 96.04 & 5.04 & 81.13 & 98.59 & 9.72 & 88.10 & 97.85 & 7.95 \\
 Sync2Gen~\cite{yang2021scene-syc2gen}  & 31.07 & 82.97 & 2.24 & 46.05 & 88.02 & 4.96 & 48.45 & 84.57 & 7.52 \\ 
 ATISS~\cite{paschalidou2021atiss}  & 18.60 & 61.71 & 0.78 & 38.66  & 71.34 & 0.64 & 40.83  & 72.66 & 0.69 \\ %\hline 
DiffuScene~\cite{tang2023diffuscene} & 18.29 & 53.52 & 0.35 & 32.60 & 55.50 & 0.22 & 36.18 & 57.81 & 0.21 \\ %
Graph-to-3D~\cite{graph23d} & 61.24  & 74.03  & 1.79  & 54.11 & 76.18 & 1.68 &41.13 & 79.37 & 2.04 \\ %\hline
Graph-to-Box~\cite{graph23d} & 55.28  & 69.48 & 1.02 & 50.29  & 73.25& 1.42 &48.77  & 78.41 & 1.81 \\  %\hline
3D-SLN~\cite{3dsln} & 58.17 & 71.27 & 1.38 &49.67 & 75.39 & 1.44 &47.29  & 76.29 & 1.77\\ %\hline
ScenePrior~\cite{nie2023learning} &   24.88    &    83.26   &   0.43  &    46.25     &    89.27  &   0.58   &     44.28      &    88.07   &   0.31   \\ %\hline 
FastSynth~\cite{fastsynth} & 31.89 & 83.61 & 2.40 & 51.26 &  90.12 & 5.26 & 57.22 & 88.21 & 6.27 \\ %
Sceneformer~\cite{wang2021sceneformer} & 33.61  & 85.38 & 1.86 & 61.08  & 85.94 & 5.18 & 63.54  & 90.20 & 3.13 \\ \hline

\text { Ours } &   \textbf{15.21}       &   \textbf{51.24}    &  \textbf{ 0.18}   &   \textbf{25.46}   &     \textbf{ 51.78}   &  \textbf{ 0.16 }  &    \textbf{31.24 }  &     \textbf{52.91}   &  \textbf{ 0.15 }  \\
\hline \hline
\end{tabular}
%\vspace{-9mm}
\end{table*}

\vspace{-2mm}
\subsection{Comparisons with State-of-the-art Methods}
%\vspace{-2mm}
\noindent \revise{\textbf{Generative Quality Evaluation:}} Table~\ref{tab:generation} compares the indoor scene generation quality of our method with existing state-of-the-art. Only our method performs (single view) sketch and knowledge guided 3D scene generation. Among the unconditional methods in Table~\ref{tab:generation}, the diffusion based DiffuScene~\cite{tang2023diffuscene} achieves better performance than Sync2Gen~\cite{yang2021scene-syc2gen}. Although graph-based methods perform well on individual object generation (Fig~\ref{fig:vis}(c)), these methods require a complete scene graph description 
that still does not specify the relative locations of objects. 
Hence, graph-based methods do not perform well in complete scene generation and deviate significantly from the target scene.
Graph-to-Box, a variant of Graph-to-3D, focuses only on learning the object layout.
Image based methods synthesize scenes under strict 2D-3D consistency. ScenePrior~\cite{nie2023learning} achieves better CKL, indicating the accuracy of object classes. 
Layout-based methods, such as ATISS~\cite{paschalidou2021atiss}, start from a given layout, often a top-down wall rendering image, and perform better in terms of generation diversity and quality. However, they do not always generate reasonable scene results. Our \names outperforms current state-of-the-art methods in quality evaluation and achieves 17.41\% FID, 3.63\% SCA, and 37.18\% CKL better than the nearest competitor DiffuScene~\cite{tang2023diffuscene} in the Dining Room category. Figure~\ref{fig:vis} shows a qualitative comparison between Sync2Gen, ATISS, Graph-to-3D, and \names.

\noindent \revise{\textbf{Sketch based Generation Evaluation:} Table~\ref{tab:sketch} compares the quality of scene generation from sketches. As there are no previous generative methods specifically designed for sketch-based generation, for a fair comparison, we adopt the current state-of-the-art methods, ATISS and Sceneformer, that allow the plugin of additional conditions. 
We append the additional sketch to the original attention modules of ATISS and Sceneformer and refer to them as ATISS-$S$ and Sceneformer-$S$, where $S$ indicates the addition of the sketch condition. We further augment these methods by concatenating the knowledge-enhanced sketch condition, resulting in the baseline models ATISS-$SK$ and Sceneformer-$SK$ respectively.
Additionally, we compare with the most relevant prior work in sketch-based scene synthesis, Sketch2Scene~\cite{xu2013sketch2scene}. Sketch2Scene optimizes scenes to closely resemble examples in a repository while adhering to constraints from input sketches. This is done through sketch-based co-retrieval and co-placement of 3D models, ensuring similarity to reference scenes while maintaining originality. In the implementation of Sketch2Scene, since our sketch is included as a whole and lacks related pixel class information, we employ DBSCAN to manually cluster the object sketches as required in the inference stage. Note that while Sketch2Scene is the most relevant prior work in sketch-based scene generation, it necessitates an additional 3D model repository. In contrast, our method generates scenes in an end-to-end manner, without the need for such repositories.}

\begin{table}[tbp]
\centering
%\footnotesize
\caption{\label{tab:sketch} Results for 3D scene generation based on a given sketch (and knowledge).}
\vspace{-4mm}
\scriptsize
%\resizebox{1\columnwidth}{!}{
\addtolength{\tabcolsep}{1pt}
\begin{tabular}{l|c|c|c|c|c|c|c|c| c}
\hline \hline \multirow{2}{*}{\text { Method }} & \multicolumn{3}{|c|}{\text { Bedroom }} & \multicolumn{3}{|c|}{\text { Dining room }} & \multicolumn{3}{|c}{\text { Living room }} \\ \cmidrule{2-10} 
& FID  $\downarrow$ & $\mathrm{KID} \downarrow$ & $\mathrm{SCA}$ \% &   $\text { FID } \downarrow$ & $\mathrm{KID} \downarrow$ & $\text { SCA } \%$ &  $\text { FID } \downarrow$ & $\mathrm{KID} \downarrow$ & $\text { SCA } \%$  \\
\hline  
%\hline 
Sceneformer-$S$~\cite{tang2023diffuscene}  & 37.21 & 13.04 & 88.37 &   64.38 & 14.21 & 87.16 &   65.78 & 15.03 & 91.14  \\
ATISS-$S$~\cite{tang2023diffuscene}  & 21.33 & 3.87 & 64.37 &   42.53 & 6.97 & 74.28 &   44.24 & 7.29 & 77.61  \\
Sceneformer-$SK$~\cite{tang2023diffuscene}  & 24.68 & 8.46 & 81.66 &   52.07 & 8.78 & 80.33 &   59.43 & 10.48 & 86.91  \\
ATISS-$SK$~\cite{tang2023diffuscene}  & 18.47& 1.58 & 58.37 &   37.24 & 2.47 & 63.05 &   39.10 & 3.08 & 63.35  \\ 
Sketch2Scene~\cite{paschalidou2021atiss}   & 22.47 & 7.18 & 55.78 &   41.35 & 6.91 & 72.38 &  58.79 & 7.27 & 84.81   \\ \hline
\text { Ours } &   \textbf{15.21}       &   \textbf{1.12}    &  \textbf{ 51.24}   &   \textbf{25.46}   &     \textbf{ 0.49}   &  \textbf{ 51.78}  &    \textbf{31.24}  &     \textbf{0.71}   &  \textbf{ 52.91 }  \\
\hline \hline
\end{tabular} %}
\vspace{-4mm}
\end{table} % with 3D-FRONT dataset

\begin{table}[h]
\centering
%\footnotesize
\vspace{-7mm}
\caption{\label{tab:completion} Results for 3D scene completion from random initial 3-5 objects.}
\vspace{-4mm}
\scriptsize
%\resizebox{1\columnwidth}{!}{
\addtolength{\tabcolsep}{0pt}
\begin{tabular}{l|c|c|c|c|c|c|c|c| c}
\hline \hline \multirow{2}{*}{\text { Method }} & \multicolumn{3}{|c|}{\text { Bedroom }} & \multicolumn{3}{|c|}{\text { Dining room }} & \multicolumn{3}{|c}{\text { Living room }} \\ \cmidrule{2-10} 
& FID  $\downarrow$ & $\mathrm{KID} \downarrow$ & $\mathrm{SCA}$ \% &   $\text { FID } \downarrow$ & $\mathrm{KID} \downarrow$ & $\text { SCA } \%$ &  $\text { FID } \downarrow$ & $\mathrm{KID} \downarrow$ & $\text { SCA } \%$  \\
\hline  ATISS~\cite{paschalidou2021atiss}   & 30.54 & 2.38 & 26.73 &   42.65 & 8.32 & 43.99 &  45.39 & 8.08 & 41.26   \\
%\hline 
DiffuScene~\cite{tang2023diffuscene}  & 27.32 & 1.92 & 40.30 &   40.99 & 6.31 & 49.06 &   43.72 & 8.37 & 46.48  \\
\hline  Ours &   \textbf{21.84 }   &   \textbf{1.58 }   &   \textbf{45.47}    &      \textbf{33.03}   &   \textbf{ 5.18}   &  \textbf{ 49.45 }   &    \textbf{35.74}  &  \textbf{ 6.51 }   &   \textbf{48.76}       \\
\hline \hline
\end{tabular} %}
\vspace{-7mm}
\end{table} 

\begin{wraptable}{r}{0.5\textwidth}
    %\footnotesize
    \vspace{-12mm}
    \caption{\label{tab:ablation} Module ablation study}% on Module validation.}
    \vspace{-1mm}
    \scriptsize
	\centering
	\addtolength{\tabcolsep}{0pt}
	%\vspace{-3mm}
	\begin{tabular}{c c | c   c   c }
		\hline \hline
	Sketch/Knowledge & SF	& FID  $\downarrow$ & $\mathrm{SCA}$ \% & $\text { CKL } \downarrow$ \\ \midrule
 $\times$/$\checkmark$  & $\checkmark$   &   32.26    &   58.31   &    0.61   \\ \midrule
   ResNet50/$\times$    &  $\checkmark$  &   34.76   &   58.94   &   1.07\\ \midrule
   ViT/$\times$         &  $\checkmark$  &  33.29    &   56.81   &  0.85    \\ \midrule
  ResNet50/$\checkmark$ &  $\checkmark$  &  24.68    &   52.70   &    0.18    \\\midrule
   ViT/$\checkmark$     & $\times$       &  25.83    &   54.19   &    0.37    \\\midrule
  ViT/$\checkmark$ & $\checkmark$   &  \textbf{ 23.97}    &   \textbf{51.97}    &  \textbf{ 0.16}     \\ % correct some typos.
\hline
		\bottomrule
		\vspace{-4mm}
	\end{tabular}
\vspace{-8mm}
\end{wraptable}

\begin{figure}[!t] 
		
		\centering
        \includegraphics[width=0.98\textwidth,clip=false]{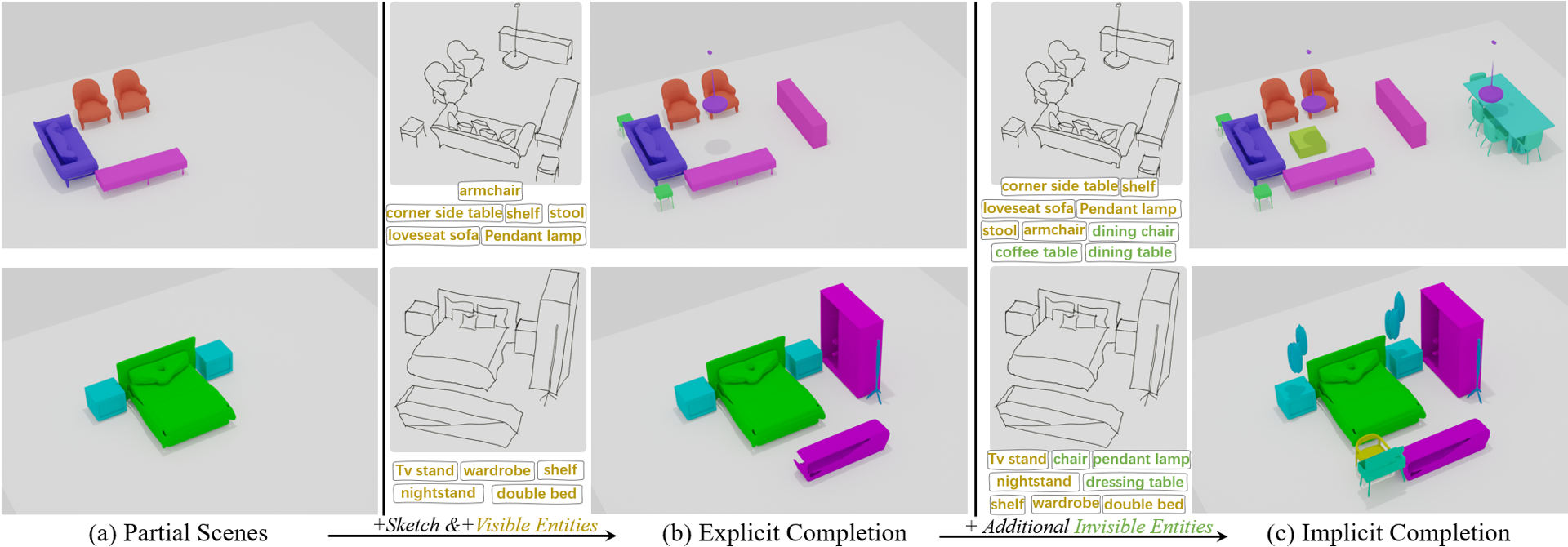}
		\vspace{-3mm}
		\caption{\label{fig:visscp} Demonstration of sketch \& knowledge guided scene completion.  In explicit completion, the sketch and user-specified entities complement each other. Beyond explicit instructions, the additional invisible entities are inferred based on knowledge and the current visible objects to generate plausible extra objects in the scene. }
		\vspace{-7mm}
	\end{figure}

\noindent \textbf{Scene Completion:} We compare against ATISS~\cite{paschalidou2021atiss} and DiffuScene~\cite{tang2023diffuscene} for scene completion. For our \names and DiffuScene, we apply the DDIM inversion process, akin to image in-painting~\cite{rombach2022high-latentdiffusion}, to the scene representation of the known furniture. A partial scene is obtained by the learned reverse chain following Eq.~\ref{eq:sample}, and is then combined with the known scene to form the completed scene. More specifically, we retrain the model by randomly masking furniture in the sketch and test the scene completion performance. Results are given in Table~\ref{tab:completion} which show that our method performs the best on all metrics achieving average 19.12\%FID, 20.06\%KID, 2.61\%SCA better than the nearest competitor DiffuScene~\cite{tang2023diffuscene}. Fig.~\ref{fig:visscp} shows qualitative results. %Our method performs plausible scene completions introducing the missing entities in the sketch.

\noindent \textbf{Ablation Study:} 
\revise{We perform ablation study on the condition modules to verify their contributions in Table~\ref{tab:ablation}.
We employ ResNet50~\cite{ResNet} and ViT~\cite{dosovitskiy2020vit} with 8 attention blocks and 8 attention heads as our sketch encoder. Sketch alone can not achieve good performance and performs worse than using only knowledge guidance. 
In the absence of knowledge, ViT shows some improvement over ResNet, but when knowledge is present, the enhancement from the sketch encoder type (ViT vs ResNet) becomes minimal. We also drop SF module for comparison (6th row).
We show the More importantly, sketch and knowledge base complement each other really well and significantly improve performance when working together to jointly promote the overall quality of generation.}

\vspace{-2mm}
\subsection{Knowledge Transfer to ScanNet}
\vspace{-1mm}

\revise{Knowledge transfer study is conduct to evaluate the effectiveness of the knowledge base across datasets.}
In our architecture, the sketch guides the spatial distribution of the objects, while the knowledge helps establish their relationships and resolves ambiguities in the sketch to generate plausible scenes. As shown in Table~\ref{tab:transfer}, we compare three baseline implementations of knowledge base: 1) Empty: We use an empty relationship knowledge base (parameters set to zero). 2) 3DFRONT: We directly use the external knowledge base constructed on 3D-FRONT for generation on ScanNet. 3) ScanNet: We re-train  
the knowledge base on ScanNet and then use it for generation on ScanNet. 
As expected, without relationship knowledge base, the results are much worse than when knowledge base is used. Interestingly, the knowledge extracted on 3DFRONT generates ScanNet scenes (row 2) as good as when knowledge is extracted from ScanNet itself to generate ScanNet scenes (row 3) with a very minor drop in performance on all metrics i.e. 0.34 FID, 0.02 KID, 0.95\%SCA, and 0.01 CKL. This shows that our constructed knowledge base effectively transfers across datasets. 

\begin{wraptable}{r}{0.4\textwidth}
%\begin{table}[t]
\centering
%\footnotesize
\vspace{-12mm}
 \caption{\label{tab:transfer} Knowledge transfer to ScanNet dataset.}% Without knowledge base (KB Empty), the performance is quite low. Directly using knowledge base (KB) constructed from 3DFRONT (row 2) to generate ScanNet scenes performs as good as when KB is constructed from the ScanNet data itself. This shows high transferability of the KB.}
\vspace{-1mm}
\scriptsize
%\resizebox{1\columnwidth}{!}{
\addtolength{\tabcolsep}{-1pt}
	\begin{tabular}{c| c   c   c  c }
		\hline\hline
		KB Source& FID  $\downarrow$ & $\mathrm{KID} \downarrow$ & $\mathrm{SCA}$ \% & $\text { CKL } \downarrow$ \\ \midrule
   Empty &  44.27     &  3.08     &  75.30     &    1.54   \\ \midrule
 3DFRONT  &  33.81    &   0.83    &   55.18    &    0.18   \\ \midrule
   ScanNet &   \textbf{33.47}    &   \textbf{0.81}    &    \textbf{54.23}   &    \textbf{0.17}   \\ \midrule
		\bottomrule
	\end{tabular}

\vspace{-6mm}
%\end{table} 
\end{wraptable}

\vspace{-2mm}
\section{Conclusion}
\vspace{-1mm}
We proposed a novel sketch based knowledge-enhanced diffusion method for generating customized, diverse, and plausible 3D scenes. Our method conditions the denoising process with a hand-drawn sketch of the required scene and cues from object relationship knowledge. Given the strong generative ability of the base diffusion model, our method can take a hand-drawn sketch along with entity information to generate diverse scenes that align well with user specifications. We  introduced a new condition for generation that incorporates external knowledge graphs, consisting of a set of well-defined relationship tuples. External knowledge helps resolve ambiguities for visible objects and their relationships in the hand-drawn sketches as well as introduce additional objects that are specified entities but not drawn in the sketch. 
Experimental results demonstrate that our model achieves state-of-the-art performance in 3D scene generation and shows promising results for the task of 3D scene completion as well.

\par\vfill\par

\section{Acknowledgement}
This research was supported by National Key R$\&$D Program of China under Grant 2023YFB4704800, National Natural Science Foundation of China under Grant 62293512, 62373293, 62293515, 62203160, and by ARC Discovery Project DP240101926. Ajmal Mian is the recipient of an ARC Future Fellowship Award (project number FT210100268) funded by the Australian Government.

%\par\vfill\par

\bibliographystyle{splncs04}
\bibliography{egbib}

\end{document}